\documentclass[runningheads]{llncs}

\usepackage{graphicx}

\usepackage{multirow}
\usepackage{adjustbox}
\usepackage{booktabs}  
\usepackage[caption=false]{subfig}
\usepackage[table]{xcolor}
\usepackage{hyperref}

\usepackage{pgfplotstable}
\usepackage{paralist}
\usepackage{bbding}

\pgfplotstableset{
    color cells/.style={
        col sep=comma,
        string type,
        postproc cell content/.code={%
                \pgfkeysalso{@cell content=\rule{0cm}{2.4ex}\cellcolor{red!##1}\pgfmathtruncatemacro\number{##1}\ifnum\number>20\color{white}\fi##1}%
                },
        columns/x/.style={
            column name={},
            postproc cell content/.code={}
        }
    }
}
\pgfplotsset{compat=1.7}

\makeatletter
\newcommand{\printfnsymbol}[1]{%
  \textsuperscript{\@fnsymbol{#1}}%
}
\makeatother

\begin{document}

\title{Reproducibility, Replicability and Beyond: Assessing Production Readiness of Aspect Based Sentiment Analysis in the Wild}
%
%

\author{Rajdeep Mukherjee\inst{1 (}\Envelope\inst{)}\thanks{Equal contribution} \and
Shreyas Shetty\inst{2}\printfnsymbol{1} \and
Subrata Chattopadhyay\inst{1} \and
Subhadeep Maji\inst{3}\thanks{Work done while at Flipkart} \and
Samik Datta\inst{3}\printfnsymbol{2} \and
Pawan Goyal\inst{1}}

\authorrunning{R. Mukherjee et al.}

\institute{Indian Institute of Technology Kharagpur, India\\
\email{rajdeep1989@iitkgp.ac.in, subrata.ctj@gmail.com, pawang@cse.iitkgp.ac.in}
\and
Flipkart Internet Private Limited, India\\
\email{shreyas.shetty@flipkart.com}
\and
Amazon India Private Limited, India\\
\email{\{subhadeepmaji, datta.samik\}@gmail.com}}

\maketitle  
\vspace{-1em}
\begin{abstract}
With the exponential growth of online marketplaces and user-generated content therein, aspect-based sentiment analysis has become more important than ever. In this work, we critically review a representative sample of the models published during the past six years through the lens of a practitioner, with an eye towards deployment in production. First, our rigorous empirical evaluation reveals poor reproducibility: an average $4-5\%$ drop in test accuracy across the sample. Second, to further bolster our confidence in empirical evaluation, we report experiments on two challenging data slices, and observe a consistent $12-55\%$ drop in accuracy. Third, we study the possibility of transfer across domains and observe that as little as $10-25\%$ of the domain-specific training dataset, when used in conjunction with datasets from other domains within the same locale, largely closes the gap between complete cross-domain and complete in-domain predictive performance. Lastly, we open-source two large-scale annotated review corpora from a large e-commerce portal in India in order to aid the study of replicability and transfer, with the hope that it will fuel further growth of the field.

\end{abstract}

\section{Introduction}
In recent times, online marketplaces of goods and services have witnessed an exponential growth in terms of consumers and producers, and have proliferated in a wide spectrum of market segments, such as e-commerce, food delivery, healthcare, ride sharing, travel and hospitality, to name a few. The Indian e-commerce market segment alone is projected to grow to $300-350$M consumers and \textdollar$100-120$B revenue by $2025$ \footnote{\href{https://www.bain.com/globalassets/noindex/2020/bain_report_how_india_shops_online.pdf}{{\em How India Shops Online} -- Flipkart and Bain \& Company.}}. In the face of ever-expanding choices, purchase decision-making is guided by the reviews and ratings: Watson et al.~\cite{doi:10.1177/0022242918805468} estimates that the average product rating is the most important factor in making purchase decisions for $60\%$ of consumers. Similarly, the academic research on Aspect Based Sentiment Analysis (ABSA) has come a long way since its humble beginning in the SemEval-$2014$ \footnote{\href{http://alt.qcri.org/semeval2014/task4/}{SemEval-$2014$ Task $4$.}}. Over the past $6$ years, the accuracy on a benchmark dataset for {\em aspect term polarity} has grown by at least $11.4\%$. We ask, is this progress enough to support the burgeoning online marketplaces?

We argue on the contrary. On one hand, industrial-strength systems need to demonstrate several traits for smooth operation and delightful consumer experience. Breck et al.~\cite{46555} articulates several essential traits and presents a rubric of evaluation. Notable traits include: (a) ``All hyperparameters have been tuned"; (b) ``A simpler model is not better"; (c) ``Training is reproducible"; and (d) ``Model quality is sufficient on important data slices". On the other hand, recent academic research in several fields has faced criticisms from within the community on similar grounds: Dhillon et al.~\cite{Dhillon2020A} points out the inadequacy of benchmark dataset and protocol for few-shot image classification; Dacrema et al.~\cite{10.1145/3298689.3347058} criticises the recent trend in recommendation systems research on the ground of lack of reproducibility and violations of (a)--(c) above; Li et al.~\cite{10.1145/3308774.3308781} criticises the recent trend in information retrieval research on similar grounds. A careful examination of the recent research we conduct in this work reveals that the field of ABSA is not free from these follies.

To this end, it is instructive to turn our attention to classic software engineering with the hope of borrowing from its proven safe development practises. Notably, Kang et al.~\cite{kang2020model} advocates the use of {\em model assertions} -- an abstraction to monitor and improve model performance during the development phase. Along similar lines, Ribeiro et al.~\cite{ribeiro2020beyond} presents a methodology of large-scale comprehensive testing for NLP, and notes its effectiveness in identifying bugs in several (commercial) NLP libraries, that would not have been discovered had we been relying solely on test set accuracy. In this work, in addition to the current practice of reporting test set accuracies, we report performance on two challenging data slices -- e.g., {\em hard set} \cite{xue-li-2018-aspect}, and, {\em contrast set} \cite{gardner-etal-2020-evaluating} -- to further bolster the comprehensiveness of empirical evaluation.

For widespread adoption, data efficiency is an important consideration in real-world deployment scenarios. As an example, a large e-commerce marketplace in India operates in tens of thousands of categories, and a typical annotation cost is $3$\textcent~per review. In this work, we introduce and open-source two additional large-scale datasets curated from product reviews in lifestyle and appliance categories to aid replicability of research and study of transfer across domains and locales (text with similar social/linguistic characteristics). In particular, we note that just a small fraction of the in-domain training dataset, mixed with existing in-locale cross-domain training datasets, guarantees comparable test set accuracies.

In summary, we make the following notable contributions:
\begin{compactitem}
    \item Perform a thorough reproducibility study of models sampled from a public leaderboard \footnote{\href{https://paperswithcode.com/sota/aspect-based-sentiment-analysis-on-semeval}{{\em Papers With Code}: ABSA on SemEval 2014 Task 4 Sub Task 2.}} that reveals a consistent $4-5 \%$ drop in reported test set accuracies, which is often larger than the gap in performance between the winner and the runner-up.
    \item Consistent with the practices developed in software engineering, we bolster the empirical evaluation rigour by introducing two challenging data slices that demonstrates an average $12-55\%$ drop in test set accuracies.
    \item We study the models from the perspective of data efficiency and note that as little as $10-25\%$ of the domain-specific training dataset, when used in conjunction with existing cross-domain datasets from within the same locale, largely closes the gap in terms of test set accuracies between complete cross-domain training and using $100\%$ of the domain-specific training instances. This observation has immense implications towards reduction of annotation cost and widespread adoption of models.
    \item We curate two additional datasets from product reviews in lifestyle and appliances categories sampled from a large e-commerce marketplace in India, and make them publicly accessible to enable the study of replicability. 
\end{compactitem}
 

\section{Desiderata and Evaluation Rubric} \label{sec:desiderata}
Reproducibility and replicability have been considered the gold-standard in academic research and has witnessed a recent resurgence in emphasis across scientific disciplines: see for e.g., McArthur et al.~\cite{doi:10.1116/1.5093621} in the context of biological sciences and Stevens et al.~\cite{10.3389/fpsyg.2017.00862} in the context of psychology. We follow the nomenclature established in \cite{10.3389/fpsyg.2017.00862} and define {\em reproducibility} as the ability to obtain same experimental results when a different analyst uses an identical experimental setup. On the other hand, {\em replicability}, is achieved when the same experimental setup is used on a different dataset to similar effect. While necessary, these two traits are far from sufficient for widespread deployment in production.

Breck et al.~\cite{46555} lists a total of $28$ traits spanning the entire development and deployment life cycle. Since our goal is only to assess the production readiness of a class of models. we decide to forego all $14$ data-, feature- and monitoring-related traits. We borrow $1$ (``Training is reproducible") and $2$ (``All hyperparameters have been tuned" and ``Model quality is sufficient on important data slices") traits from the infrastructure- and modeling-related rubrics, respectively.

Further, we note that the ability to transfer across domains/locales is a desirable trait, given the variety of market segments and the geographic span of online marketplaces. In other words, this expresses data efficiency and has implications towards lowering the annotation cost and associated deployment hurdles. Given the desiderata, we articulate our production readiness rubric as follows:
\begin{compactitem}
    \item {\em Reproducibility.} A sound experimental protocol that minimises variability across runs and avoids common pitfalls (e.g., hyperparameter-tuning on the test dataset itself) should reproduce the reported test set accuracy within a reasonable tolerance, not exceeding the reported performance gap between the winner and the runner-up in a leaderboard. \S \ref{sec:experiment-protocol} articulates the proposed experimental protocol and \S \ref{sec:result} summarises the ensuing observations.
    \item {\em Replicability.} The aforementioned experimental protocol, when applied to a different dataset, should not dramatically alter the conclusions drawn from the original experiment; specifically, it should not alter the relative positions within the leaderboard. \S \ref{sec:dataset} details two new datasets we contribute in order to aid the study of replicability, whereas \S \ref{sec:result} contains the ensuing observations.
    \item {\em Performance.} Besides overall test-set accuracy, an algorithm should excel at challenging data slices such as hard- \cite{xue-li-2018-aspect} and contrast sets \cite{gardner-etal-2020-evaluating}. \S \ref{sec:result} summarises our findings when this checklist is adopted as a standard reporting practice.
    \item {\em Transferability.} An algorithm must transfer gracefully across domains within the same locale, i.e. textual data with similar social/linguistic characteristics. We measure it by varying the percentage of in-domain training instances from $0\%$ to $100\%$ and locating the inflection point in test set accuracies. See \S \ref{sec:result} for additional details.
\end{compactitem}

Note that apart from the ``The model is debuggable" and ``A simpler model is not better" traits, the remaining traits as defined by Breck et al.\cite{46555} are independent of the choice of the algorithm and is solely a property of the underlying system that embodies it, which is beyond the scope of the present study. Unlike \cite{46555}, we refrain from developing a numerical scoring system.

\section{Related Work} \label{sec:survey}
First popularised in the SemEval-$2014$ Task $4$~\cite{pontiki-etal-2014-semeval}, ABSA has enjoyed immense attention from both academic and industrial research communities. Over the past $6$ years, according to the cited literature on a public leaderboard \footnote{\href{https://paperswithcode.com/sota/aspect-based-sentiment-analysis-on-semeval}{{\em Papers With Code}: ABSA on SemEval 2014 Task 4 Sub Task 2.}}, the performance for the subtask of {\em Aspect Term Polarity} has increased from $70.48\%$ in Pontiki et al.~\cite{pontiki-etal-2014-semeval}, corresponding to the winning entry, to $82.29\%$ in Yang et al.~\cite{yang2019multi} on the laptop review corpus. The restaurant review corpus has witnessed a similar boost in performance: from $80.95\%$ in~\cite{pontiki-etal-2014-semeval} to $90.18\%$ in~\cite{yang2019multi}. 

Not surprisingly, the field has witnessed a phase change in terms of the methodology: custom feature engineering and ensembles that frequented earlier~\cite{pontiki-etal-2014-semeval} gave way to neural networks of ever-increasing complexity. Apart from this macro-trend, we notice several micro-trends in the literature: the year $2015$ witnessed a proliferation of LSTM and its variants~\cite{tang-etal-2016-effective}; years $2016$ and $2017$ respectively witnessed the introduction~\cite{tang-etal-2016-aspect} and proliferation~\cite{tay2017dyadic,ijcai2017-568,Cheng:2017:ASC:3132847.3133037,chen-etal-2017-recurrent-attention} of memory networks and associated attention mechanisms; in $2018$ research focused on CNN~\cite{xue-li-2018-aspect}, transfer learning~\cite{DBLP:journals/corr/abs-1811-10999} and transformers~\cite{li-etal-2018-transformation}, while memory networks and attention mechanisms remained in spotlight~\cite{li-etal-2018-hierarchical,wang2018learning,Huang2018AspectLS,Liu:2018:CAM:3178876.3186001}; transformer and BERT-based models prevailed in $2019$~\cite{xu-etal-2019-bert,Zeng2019LCFAL}, while attention mechanisms continued to remain mainstream~\cite{song2019attentional}.

While these developments appear to have pushed the envelope of performance, the field has been fraught with ``winner's curse" \cite{sculley2018winner}. In addition to the replicability and reproducibility crises~\cite{doi:10.1116/1.5093621,10.3389/fpsyg.2017.00862}, criticisms around inadequacy of baseline and unjustified complexity~\cite{10.1145/3298689.3347058,Dhillon2020A,10.1145/3308774.3308781} applies to this field as well. The practice of reporting performance in challenging data slices \cite{xue-li-2018-aspect} has not been adopted uniformly, despite its importance to production readiness assessment \cite{46555}. Similarly, the study of transferability and replicability has only been sporadically performed: e.g., Hu et al.~\cite{hu2019domain} uses a dataset curated from Twitter along with the ones introduced in Pontiki et al.~\cite{pontiki-etal-2014-semeval} for studying cross-domain transferability.

\section{Dataset} \label{sec:dataset}
For the {\sl Reproducibility} rubric, we consider the datasets released as part of the SemEval 2014 Task 4 - Aspect Based Sentiment Analysis \footnote{SemEval 2014: Task 4 http://alt.qcri.org/semeval2014/task4/
} for our experiments, specifically the Subtask 2 - Aspect term Polarity. The datasets come from two domains -- Laptop and Restaurant. We use their versions made available in this Github \footnote{https://github.com/songyouwei/ABSA-PyTorch} repository which forms the basis of our experimental setup.

The guidelines used for annotating the datasets were released as part of the challenge. For the {\sl Replicability} rubric, we tagged two new datasets from the e-commerce domain viz., Men's T-shirt and Television, using similar guidelines. 

The statistics for these four datasets are presented in Table \ref{tab:dataset_stats}. As we can observe, the sizes of the Men's T-shirt and Television datasets are comparable to the laptop and restaurant datasets, respectively. 

\begin{table}[h]
\centering
\caption{Statistics of the datasets showing the no. of sentences with corresponding sentiment polarities of constituent aspect terms.}
\label{tab:dataset_stats}
\begin{adjustbox}{max width=\textwidth}
\begin{tabular}{|c|c|c|c|c|c|c|c|c|}
\hline
\multirow{2}{*}{\textbf{Dataset}} & \multicolumn{4}{c|}{\textbf{Train}} & \multicolumn{4}{c|}{\textbf{Test}} \\ 
\cline{2-9} & \textbf{Positive} & \textbf{Negative} & \textbf{Neutral} & \textbf{Total} & \textbf{Positive} & \textbf{Negative} & \textbf{Neutral} & \textbf{Total} \\ 
\hline
\textbf{Laptop} & 994 & 870 & 464 & 2328 & 341 & 128 & 169 & 638 \\
\hline
\textbf{Restaurant} & 2164 & 807 & 637 & 3608 & 728 & 196 & 196 & 1120 \\ 
\hline
\textbf{Men's T-shirt} & 1122 & 699 & 50 & 1871 & 270 & 186 & 16 & 472 \\ 
\hline
\textbf{Television} & 2540 & 919 & 287 & 3746 & 618 & 257 & 67 & 942 \\ 
\hline
\end{tabular}
\end{adjustbox}
\end{table}

For the {\sl Performance} rubric, we evaluate and compare the models on two challenging subsets viz., {\em hard} as defined by Xue et al.~\cite{xue-li-2018-aspect} and {\em contrast} as defined by Gardner et al.~\cite{gardner-etal-2020-evaluating}. We describe below the process to obtain these datasets:

\begin{compactitem}
    \item \textbf{Hard data slice:} Hard examples have been defined in Xue et al.~\cite{xue-li-2018-aspect} as the subset of review sentences containing multiple aspects with different corresponding sentiment polarities. The number of such hard examples from each of the datasets are listed in Table \ref{tab:hard_set_stats}. 
    \item \textbf{Contrast data slice:} In order to create additional test examples, Gardner et al.~\cite{gardner-etal-2020-evaluating} adds perturbations to the test set, by modifying only a couple of words to flip the sentiment corresponding to the aspect under consideration. For e.g., consider the review sentence: ``I was happy with their service and food". If we change the word ``happy" with ``dissatisfied", the sentiment corresponding to the aspect ``food" changes from positive to negative. We take a random sample of $30$ examples from each of the datasets and add similar perturbations as above to create $30$ additional examples. These $60$ examples for each of the four datasets thus serve as our contrast test sets.
\end{compactitem}

\begin{table}[t]
\centering
\caption{Statistics of the Hard test sets}
\label{tab:hard_set_stats}
\begin{tabular}{|c|c|c|c|c|}
\hline
\textbf{Dataset} & \textbf{Positive} & \textbf{Negative} & \textbf{Neutral} & \textbf{Total (\% of Test Set)} \\ 
\hline
\textbf{Laptop} & 31 & 24 & 46 & 101 (15.8 \%) \\
\hline
\textbf{Restaurants} & 81 & 60 & 83 & 224 (20.0 \%) \\
\hline
\textbf{Men's T-shirt} & 23 & 24 & 1 & 48 (10.2 \%) \\ 
\hline
\textbf{Television} & 43 & 40 & 19 & 102 (10.8 \%) \\ 
\hline
\end{tabular}
\end{table}

\section{Models Compared} \label{sec:model-list}
As part of our evaluation, we focus on two families of models which cover the major trends in the ABSA research community: (i) memory network based, and (ii) BERT based. Among the initial set of models for the SemEval 14 challenge, memory network based models had much fewer parameters compared to LSTM based approaches and performed comparatively better. With the introduction of BERT~\cite{devlin-etal-2019-bert}, work in NLP has focused on leveraging BERT based architectures for a wide spectrum of tasks. In the ABSA literature, the leaderboard \footnote{https://paperswithcode.com/sota/aspect-based-sentiment-analysis-on-semeval} has been dominated by BERT based models, which have orders of magnitude more parameters than memory network based models. However, due to pre-training on large corpora, BERT models are still very data efficient in terms of number of labelled examples required. We chose three representative models from each family for our experiments and briefly describe them below:

\begin{compactitem}
    \item \textbf{ATAE-LSTM} \cite{wang-etal-2016-attention} represents aspects using target embeddings and models the context words using an LSTM. The context word representations and target embeddings are concatenated and combined using an attention layer.
    \item \textbf{Recurrent Attention on Memory (RAM)} \cite{chen-etal-2017-recurrent-attention} represents the input review sentence using a memory network, and the memory cells are weighted using the distance from the target word. The aspect representation is then used to compute attention scores on the input memory, and the attention weighted memory is refined iteratively using a GRU (recurrent) network.
    \item \textbf{Interactive Attention Networks (IAN)} \cite{10.5555/3171837.3171854} uses separate components for computing representations for both the target (aspect) and the context words. The representations are pooled and then used to compute an attention score on each other. Finally the individual attention weighted representations are concatenated to obtain the final representation for the 3-way classification task, with {\em positive}, {\em negative}, and {\em neutral} being the three classes.
    \item \textbf{BERT-SPC}~\cite{devlin-etal-2019-bert} is a baseline BERT model that uses ``[CLS] + context + [SEP] + target + [SEP]" as input for the sentence pair classification task, where `[CLS]' and `[SEP]' represent the tokens corresponding to {\em classification} and {\em separator} symbols respectively, as defined in Devlin et al.~\cite{devlin-etal-2019-bert} .
    \item \textbf{BERT-AEN} \cite{song2019attentional} uses an attentional encoder network to model the semantic interaction between the context and the target words. Its loss function uses a label smoothing regularization to avoid overfitting.
    \item \textbf{The Local Context Focus (LCF-BERT)}~\cite{Zeng2019LCFAL} is based on Multi-head Self-Attention (MHSA). It uses Context features Dynamic Mask (CDM) and Context features Dynamic Weighted (CDW) layers to focus more on the local context words. A BERT-shared layer is adopted to LCF design to capture internal long-term dependencies of local and global context.
\end{compactitem}

\section{Experimental Setup} \label{sec:experiment-protocol}
We present an extensive evaluation of the aforementioned models across the four datasets: Laptops, Restaurants, Men's T-shirt and Television, as per the production readiness rubrics defined in \S \ref{sec:desiderata}. While trying to reproduce the reported results for the models, we faced two major issues; (i) the official implementations were not readily available, and (ii) the exact hyperparameter configurations were not always specified in the corresponding paper(s). In order to address the first, our experimental setup is based on a community designed implementation of recent papers available on GitHub \footnote{https://github.com/songyouwei/ABSA-PyTorch}. Our choice for this public repository is guided by its thoroughness and ease of experimentation. As an additional social validation, the repository had 1.1k stars and 351 forks on GitHub at the time of writing. For addressing the second concern, we consider the following options; (a) use commonly accepted default parameters (for e.g., using a learning rate of $1e^{-4}$ for Adam optimizer). (b) use the public implementations to guide the choice of hyperparameters. The exact hyperparameter settings used in our experiments are documented and made available with our supporting code repository \footnote{https://github.com/rajdeep345/ABSA-Reproducibility} for further reproducibility and replicability of results. 

From the corresponding experimental protocols described in the original paper(s), we were not sure if the final numbers reported were based on the training epoch that gave the best performance on the test set, or whether the hyperparameters were tuned on a separate held-out set. Therefore, we use the following two configurations; (i) the test set is itself used as the held out set, and the model used for reporting the results is chosen corresponding to the training epoch with best performance on the test set; and (ii) 15\% of the training data is set aside as a held out set for tuning the hyperparameters and the optimal training epoch is decided corresponding to the best performance on the held out set. Finally the model is re-trained, this time with all the training data (including 15\% held out set), for the optimal no. of epochs before evaluating the test set. For both the cases, we report mean scores over $5$ runs of our experiments.

\section{Results and Discussion: Production Readiness Rubrics} \label{sec:result}

\subsection{Reproducibility and Replicability}
\begin{table*}
\centering
\caption{Performance of the models on the four datasets. The first two dataset correspond to the reproducibility study, while the next two datasets correspond to the replicability study. Towards performance study, results on the hard and contrast data slices are respectively enclosed in brackets in the last two columns. All the reproduced and replicated results are averaged across 5 runs.}
\label{tab:reproducibility_study}
\subfloat[Laptop\label{tab:laptop_results}]{
    \centering
    \begin{adjustbox}{max width=\textwidth}
    \begin{tabular}{|c|c|c|l|l|l|l|}
    \hline
    \multirow{2}{*}{\textbf{Model}} & \multicolumn{2}{c|}{\textbf{Reported}} & \multicolumn{2}{c|}{\textbf{Reproduced}} & \multicolumn{2}{c|}{\textbf{Reproduced using 15\% held out set}} \\ 
    \cline{2-7} & \multicolumn{1}{c|}{\textbf{Accuracy}} & \multicolumn{1}{c|}{\textbf{Macro-F1}} & \multicolumn{1}{c|}{\textbf{Accuracy}} & \multicolumn{1}{c|}{\textbf{Macro-F1}} & \multicolumn{1}{c|}{\textbf{Accuracy}} & \multicolumn{1}{c|}{\textbf{Macro-F1}} \\ 
    \hline
    \textbf{ATAE-LSTM}  & 68.70 & - & 60.28 & 44.33 & 58.62 (33.47, 26.00) & 43.27 (29.01, 22.00) \\ 
    \hline
    \textbf{RAM} & 74.49 & 71.35 & 72.82 & 68.34 & 70.97 (56.04, 46.00) & 65.31 (55.81, 43.16) \\ 
    \hline
    \textbf{IAN} & 72.10 & - & 69.94 & 62.84 & 69.40 (48.91, 34.67) & 61.98 (48.75, 33.40) \\ 
    \hline
    \textbf{BERT-SPC} & 78.99 & 75.03 & 78.72 & 74.52 & 77.24 (59.21, 52.00) & 72.80 (59.44, 48.67) \\ 
    \hline
    \textbf{BERT-AEN} & 79.93 & 76.31 & 78.65 & 74.26 & 75.71 (46.53, 37.33) & 70.02 (45.22, 36.20) \\ 
    \hline
    \textbf{LCF-BERT} & 77.31 & 75.58 & 79.75 & 76.10 & 77.27 (62.57, 54.67) & 72.86 (62.71, 49.56) \\ 
    \hline
    \end{tabular}
    \end{adjustbox}
}
\newline
\subfloat[Restaurant\label{tab:restaurant_results}]{
    \centering
    \begin{adjustbox}{max width=\textwidth}
    \begin{tabular}{|c|c|c|l|l|l|l|}
    \hline
    \multirow{2}{*}{\textbf{Model}} & \multicolumn{2}{c|}{\textbf{Reported}} & \multicolumn{2}{c|}{\textbf{Reproduced}} & \multicolumn{2}{c|}{\textbf{Reproduced using 15\% held out set}} \\ 
    \cline{2-7} & \multicolumn{1}{c|}{\textbf{Accuracy}} & \multicolumn{1}{c|}{\textbf{Macro-F1}} & \multicolumn{1}{c|}{\textbf{Accuracy}} & \multicolumn{1}{c|}{\textbf{Macro-F1}} & \multicolumn{1}{c|}{\textbf{Accuracy}} & \multicolumn{1}{c|}{\textbf{Macro-F1}} \\ 
    \hline
    \textbf{ATAE-LSTM} & 77.20 & - & 73.71 & 55.87 & 73.29 (52.41, 38.71) & 54.59 (47.35, 33.13) \\ 
    \hline
    \textbf{RAM} & 80.23 & 70.80 & 78.21 & 65.94 & 76.36 (59.29, 56.77) & 63.15 (56.36, 56.12) \\ 
    \hline
    \textbf{IAN} & 78.60 & - & 76.80 & 64.24 & 76.52 (57.05, 50.32) & 63.84 (55.11, 48.19) \\ 
    \hline
    \textbf{BERT-SPC} & 84.46 & 76.98 & 85.04 & 78.02 & 84.23 (68.84, 57.42) & 76.28 (68.11, 57.23) \\ 
    \hline
    \textbf{BERT-AEN} & 83.12 & 73.76 & 81.73 & 71.24 & 80.07 (51.70, 45.81) & 69.80 (48.97, 46.88) \\ 
    \hline
    \textbf{LCF-BERT} & 87.14 & 81.74 & 85.94 & 78.97 & 84.20 (69.38, 56.77) & 76.28 (69.64, 57.81) \\ 
    \hline
    \end{tabular}
    \end{adjustbox}
}
\newline
\subfloat[Men's T-shirt\label{tab:menstshirt_results}]{
    \centering
    \begin{adjustbox}{max width=\textwidth}
    \begin{tabular}{|c|l|l|l|l|}
    \hline
    \multirow{2}{*}{\textbf{Model}} & \multicolumn{2}{c|}{\textbf{Replicated}} & \multicolumn{2}{c|}{\textbf{Replicated using 15\% held out set}} \\
    \cline{2-5} & \multicolumn{1}{c|}{\textbf{Accuracy}} & \multicolumn{1}{c|}{\textbf{Macro-F1}} & \multicolumn{1}{c|}{\textbf{Accuracy}} & \multicolumn{1}{c|}{\textbf{Macro-F1}} \\ 
    \hline
    \textbf{ATAE-LSTM} & 83.13 & 55.98 & 81.65 (58.33, 40.67) & 54.84 (39.25, 30.54) \\ \hline
    \textbf{RAM} & 90.51 & 61.93 & 88.26 (83.33, 46.00) & 59.67 (56.01, 33.85) \\ 
    \hline
    \textbf{IAN} & 87.58 & 59.16 & 87.41 (63.75, 42.67) & 58.97 (42.85, 31.94) \\ 
    \hline
    \textbf{BERT-SPC} & 93.13 & 73.86 & 92.42 (89.58, 66.00) & 73.83 (60.62, 56.90) \\ 
    \hline
    \textbf{BERT-AEN} & 88.69 & 72.25 & 87.54 (50.42, 58.67) & 59.14 (32.96, 43.00) \\ 
    \hline
    \textbf{LCF-BERT} & 93.35 & 72.19 & 91.99 (91.67, 71.33) & 72.13 (62.30, 59.70) \\ 
    \hline
    \end{tabular}
    \end{adjustbox}
}
\newline
\subfloat[Television\label{tab:television_results}]{
    \centering
    \begin{adjustbox}{max width=\textwidth}
    \begin{tabular}{|c|l|l|l|l|}
    \hline
    \multirow{2}{*}{\textbf{Model}} & \multicolumn{2}{c|}{\textbf{Replicated}} & \multicolumn{2}{c|}{\textbf{Replicated using 15\% held out set}} \\
    \cline{2-5} & \multicolumn{1}{c|}{\textbf{Accuracy}} & \multicolumn{1}{c|}{\textbf{Macro-F1}} & \multicolumn{1}{c|}{\textbf{Accuracy}} & \multicolumn{1}{c|}{\textbf{Macro-F1}} \\ 
    \hline
    \textbf{ATAE-LSTM} & 81.10 & 53.71 & 79.68 (53.92, 25.33) & 52.78 (39.13, 16.80) \\ \hline
    \textbf{RAM} & 84.29 & 58.68 & 83.02 (64.31, 53.33) & 58.50 (50.07, 45.51) \\ 
    \hline
    \textbf{IAN} & 82.42 & 57.15 & 80.49 (54.31, 32.00) & 56.78 (41.67, 25.16) \\ 
    \hline
    \textbf{BERT-SPC} & 89.96 & 74.68 & 88.56 (80.20, 62.67) & 74.81 (74.32, 60.25) \\ 
    \hline
    \textbf{BERT-AEN} & 87.09 & 67.92 & 85.94 (50.39, 50.66) & 65.65 (38.08, 45.75) \\ 
    \hline
    \textbf{LCF-BERT} & 90.36 & 76.01 & 90.00 (80.98, 66.67) & 75.86 (73.72, 64.15) \\ 
    \hline
    \end{tabular}
    \end{adjustbox}
}
\end{table*}
Tables 3(a) and 3(b) show our {\sl reproducibility} study for the Laptop and Restaurant datasets, respectively. For both the datasets, we notice a consistent 1-2\% drop in accuracy and macro-f1 scores when we try to reproduce the reported numbers in the corresponding papers. Only exceptions were LCF-BERT for Laptop and BERT-SPC for Restaurant dataset, where we got higher numbers than the reported ones. For ATAE-LSTM, the drop observed was much larger than other models. We notice an additional 1-2\% drop in accuracy when we use 15\% of the training set as a held-out set to pick the best model. These numbers indicate that the actual performance of the models is likely to be slightly worse than what is quoted in the papers, and the drop sometimes is larger than the difference between the performance of two consecutive methods on the leaderboard.

To study the {\sl replicability}, Tables 3(c) and 3(d) summarise the performance of the individual models on the Men's T-shirt and Television datasets, respectively. We introduce these datasets for the first time and report the performance of all $6$ models under the two defined configurations: test set as held out set, and 15\% of train set used as held out set. We notice a similar drop in performance when we follow the correct experimental procedure (hyperparameter tuning on 15\% train data as held-out set). Therefore, following a consistent and rigorous experimental protocol helps us to get a better sense of the true model performance.

\subsection{Performance on the Hard and Contrast data slices}
As per the {\sl performance} rubric, we investigate the performance of all $6$ models on both {\em hard} and {\em contrast} test sets, using the correct experimental setting (15\% train data as held out set). The results are shown in brackets (in same order) in the last two columns of Tables 3(a), 3(b), 3(c), and 3(d) for the four datasets, respectively. We observe a large drop in performance on both these challenging data slices across models. LCF-BERT consistently performs very well on these test sets. Among memory network based models, RAM performs the best.

\subsection{Transferability rubric: Cross domain experiments}
In a production readiness setting, it is very likely that we will not have enough labelled data across individual categories and hence it is important to understand how well the models are able to transfer across domains. To understand the transferability of models across datasets, we first experiment with cross domain combinations. For each experiment, we fix the test set (for e.g., Laptop) and train three separate models, each with one of the other three datasets as training sets (Restaurant, Men's T-shirt, and Television in this case). Consistent with our experimental settings, for each such combination, we use 15\% of the cross-domain data as held-out set for hyperparameter tuning, re-train the corresponding models with all the cross-domain data and obtain the scores for the in-domain set (here Laptop) averaged across $5$ different runs of the experiment.

\begin{table}[t]
    \centering
    \caption{Transferability: Average drop between in-domain and cross-domain accuracies for each dataset pair for (a) BERT based and (b) Memory network based models. Rows correspond to the train set. Columns correspond to the test set.}  
    \label{tab:transferability}
    \begin{tabular}{cc}
        \includegraphics[width=0.49\textwidth]{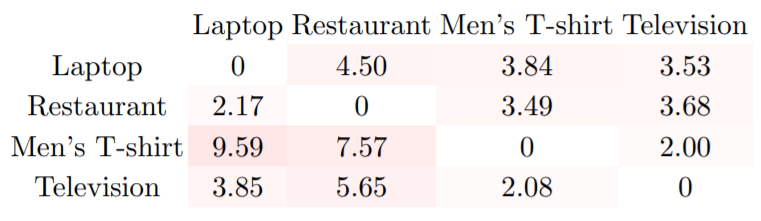} & \includegraphics[width=0.49\textwidth]{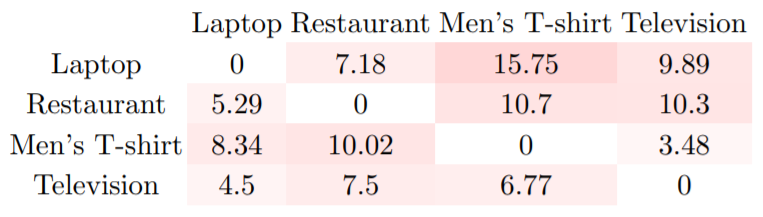} \\
        (a) BERT based models & (b) Memory network based models
    \end{tabular}
    \label{tab:my_label}
\end{table}

    

Table \ref{tab:transferability} summarises the results averaged across the BERT-based models and Memory network based models, respectively on the four datasets. The rows and columns correspond to the train and test sets, respectively. The diagonals correspond to the in-domain experiments (denoted by $0$) and each off-diagonal entry denotes the average drop in model performance for the cross-domain setting compared to the in-domain combination. 

From Table \ref{tab:transferability} we observe that on an average the models are able to generalize well across the following combinations, which correspond to a lower drop in the cross domain experiments: (i) Laptops and Restaurants, and (ii) Men's T-shirt and Television. For instance, when testing on the Restaurant dataset, BERT based and memory network based models respectively show an average of $\sim$4 and $\sim$7 point absolute drops in \% accuracies, when trained using the Laptop dataset. The drops are higher for the other two training sets. Interestingly, the generalization is more pronounced across locales rather than domains, contrary to what one would have expected. For e.g., we notice better transfer from Men's T-shirt $\rightarrow$ Television (similarity in locale) than in the expected Laptop $\rightarrow$ Television (similarity in domain). Given that our task is that of detecting sentiment polarities of aspect terms, this observation might be attributed to the similarity in social/linguistic characteristics of reviews from the same locale.

Further, in the spirit of {\sl transferability}, we consider the closely related locales as identified above -- \{Laptop, Restaurant\} and \{Men's T-shirt, Television\}, and conduct experiments to understand the incremental benefits of adding in-domain data on top of cross domain data, i.e., what fraction of the in-domain training instances can help to cover the gap between purely in-domain and purely cross-domain performance largely. For each test dataset, we take examples from the corresponding cross-domain dataset in the same locale as training set and incrementally add in-domain (10\%, 25\% and 50\%) examples to evaluate the performance of the models. Table \ref{tab:incremental} summarises the results from these experiments for the BERT based models (a) and memory network based models (b). For instance, on the Restaurant dataset, the average cross-domain performance (i.e., trained on Laptop) across the three BERT-based models is 78.3 (first row), while the purely in-domain performance is 82.8 (last row). We observe that among all increments, adding 10\% of the in-domain dataset (second row) gives the maximum improvement, and is accordingly defined as the inflection point, which is marked in bold. In Table \ref{tab:incremental} (a), we report the accuracy scores (averaged over 5 runs) for the individual BERT based models (BERT-AEN, BERT-SPC, LCF-BERT) in brackets, in addition to the average numbers. As we can see, the variability in the numbers across models is low. For the memory network based models, on the other hand, the variability is not so low, and the corresponding scores have been shown in Table \ref{tab:incremental} (b) in the order (ATAE-LSTM, IAN, RAM).

Interestingly, we notice that in most of the cases, the inflection point is obtained upon adding just 10\% in-domain examples and the model performance reaches within $0.5-2$\% of purely in-domain performance, as shown in Table \ref{tab:comparison}. While in a few cases, it happens by adding 25-50\% in-domain samples. This is especially useful from the production readiness perspective since considerably good performance can be achieved by using limited in-domain labelled data on top of cross-domain annotated data from the same locale.

\begin{table}[t]
    \caption{Transferability: Results on including incremental in-domain training data. The rows correspond to cross-domain performance (0), adding 10\%, 25\% and 50\% in-domain dataset to the cross-domain. To improve illustration, we repeat in-domain results. Inflection points for each dataset are boldfaced.}
    \label{tab:incremental}
    \subfloat[Variance across BERT based models (BERT-AEN, BERT-SPC, LCF-BERT) is small.]{
        \centering
        \begin{adjustbox}{max width=\textwidth}
        \begin{tabular}{l|c|c|c|c}
        \toprule
        \textbf{\% in-domain} & \textbf{Laptop} & \textbf{Restaurant} & \textbf{Men's T-shirt} & \textbf{Television} \\
        \midrule   
        \textbf{0} & 74.6 (73.6, 74.6, 75.5) & 78.3 (77.3, 77.8, 79.9) & 88.6 (86.3, 89.6, 89.8) & 86.1 (83.5, 87.5, 87.4) \\
        \textbf{10} & \textbf{76.5} (73.9, 76.6, 78.9) & \textbf{81.6} (80.1, 81.5, 83.3) & 88.9 (85.7, 90.6, 90.4) & 83.8 (82.0, 86.1, 83.2)  \\
        \textbf{25} & 76.3 (74.8, 77.0, 77.0) & 82.1 (79.8, 82.8, 83.7) & \textbf{90.0} (87.2, 91.7, 91.0) & 86.3 (83.8, 86.8, 88.2)  \\
        \textbf{50} &  78.2 (76.4, 79.2, 78.9) & 82.9 (80.8, 83.6, 84.4) & 90.1 (86.8, 91.3, 92.3) & \textbf{87.2} (85.5, 88.2, 87.8) \\
        \midrule
        \textit{In-domain} & 76.7 (75.7, 77.2, 77.3) & 82.8 (80.1, 84.2, 84.2) & 90.6 (87.5, 92.4, 92.0) & 88.2 (85.9, 88.6, 90.0) \\
        \bottomrule
        \end{tabular}
        \end{adjustbox}
    }
    \newline
    \subfloat[Variance across Memory network models (ATAE-LSTM, IAN, RAM) is significant.]{
        \centering
        \begin{adjustbox}{max width=\textwidth}
        \begin{tabular}{l|c|c|c|c}
        \toprule
        \textbf{\% in-domain} & \textbf{Laptop} & \textbf{Restaurant} & \textbf{Men's T-shirt} & \textbf{Television}\\
        \midrule   
        \textbf{0} & 61.0 (58.6, 60.9, 63.6) & 68.2 (68.3, 68.0, 68.3) & 79.0 (76.6, 78.6, 81.9) & 77.6 (75.4, 77.8, 79.6) \\
        \textbf{10} & \textbf{65.1} (60.7, 65.6, 69.1) & \textbf{73.0} (70.1, 74.1, 74.9) & \textbf{83.8} (80.3, 84.1, 86.9) & \textbf{79.1} (77.1, 79.1, 81.2) \\
        \textbf{25} & 65.3 (59.9, 66.2, 69.8) & 74.8 (72.2, 75.4, 76.6) & 85.1 (82.9, 86.0, 86.4) & 80.0 (78.7, 79.8, 81.5) \\
        \textbf{50} & 66.2 (60.5, 68.7, 69.5) & 75.0 (72.9, 75.3, 76.8) & 85.8 (82.7, 86.1, 88.4) & 80.6 (78.8, 80.7, 82.4)\\
        \midrule
        \textit{In-domain} & 66.3 (58.6, 69.4, 71.0) & 75.4 (73.3, 76.5, 76.4) & 85.8 (81.7, 87.4, 88.3) & 81.1 (79.7, 80.5, 83.0) \\
        \bottomrule
        \end{tabular}
        \end{adjustbox}
    }
\end{table}

\begin{table}
  \centering
  \caption{Performance scorecard in accordance with the rubric: {\em reproducibility} -- \% drop in test set accuracy across Laptop and Restaurant, resp.; {\em replicability} -- rank in leaderboard for Men's T-shirt and Television, resp. (rank obtained from avg. test set accuracy on Laptop and Restaurant); {\em performance} -- \% drop in test set accuracy (averaged across all four datasets) with hard and contrast-set data slices, resp.; {\em transferability} -- \% drop in test set accuracy in cross-domain setting, and upon adding in-domain training instances as per the inflection point, resp. (averaged over the four datasets)}
    \label{tab:comparison}
    \begin{tabular}{l|*{4}c}
    \toprule
    \textbf{Model} & \textbf{Reproducibility} & \textbf{Replicability} & \textbf{Performance} & \textbf{Transferability} \\
    \midrule
    \textbf{ATAE-LSTM} & (14.67, 5.06) & 6, 6 (6) &  (33.07, 55.31)  & (4.60, 1.44) \\
    \textbf{RAM} & (4.73, 4.82) & 3, 4 (4) & (17.88, 36.12) & (8.06, 2.06) \\
    \textbf{IAN} & (3.74, 2.64) & 5, 5 (5) & (28.64, 48.93) & (9.22, 3.55) \\
    \textbf{BERT-SPC} & (2.22, 0.27) & 1, 2 (2) & (13.53, 30.58) & (3.83, 1.33) \\
    \textbf{BERT-AEN} & (5.28, 3.67) & 4, 3 (3) & (39.44, 41.88) & (2.61, 0.83) \\
    \textbf{LCF-BERT} & (0.05, 3.37) & 2, 1 (1) &  (11.75, 27.55) & (3.14, 0.64) \\
    \bottomrule
    \end{tabular}
\end{table}

\subsection{Summary comparison of the different models under the production readiness rubrics}
We now make an overall comparison across different models considered in this study under our production readiness rubrics. Table \ref{tab:comparison} shows the various numbers across these rubrics. Under {\sl reproducibility}, we observe a consistent drop in performance even for the BERT-based models, atleast for one of the two datasets, viz. Laptop and Restaurant. For Memory network based models, while there is a considerable drop across both the datasets, the drop for the Laptop dataset is quite noteworthy. Under {\sl replicability}, we observe that the relative rankings of the considered models remain quite stable for the two new datasets, which is a good sign. Under {\sl performance}, we note a large drop in test set accuracies for all the models across the two challenging data slices, with a minimum drop of 11-27\% for LCF-BERT. Surprisingly, BERT-AEN suffered a huge drop in performance for both hard as well as contrast data slices. This is a serious concern and further investigation is needed to identify the issues responsible for this significant drop. Under {\sl transferability}, while there is consistent drop in cross-domain scenario, the drop with the inflection point, corresponding to a meager addition of 10-25\% of in-domain data samples, is much smaller.

\subsection{Limitations of the present study}
While representative of the modern trend in architecture research, memory network- and BERT-based models do not cover the entire spectrum of the ABSA literature. Important practical considerations, such as debuggability, simplicity and computational efficiency, have not been incorporated into the rubric. Lastly, a numeric scoring system based on the rubric would have made its interpretation objective. We leave them for a future work.

\section{Conclusion}
Despite the limitations, the present study takes an important stride towards closing the gap between empirical academic research and its widespread adoption and deployment in production. In addition to further strengthening the rubric and judging a broader cross-section of published ABSA models in its light, we envision to replicate such study in other important NLP tasks. We hope the two contributed datasets, along with the open-source evaluation framework, shall fuel further rigorous empirical research in ABSA. We make all the codes and datasets publicly available \footnote{https://github.com/rajdeep345/ABSA-Reproducibility}.

\bibliographystyle{splncs04}
\bibliography{main}

\end{document}